%% file: main.tex
\documentclass[conference]{IEEEtran}
\IEEEoverridecommandlockouts

\usepackage{amsmath,amssymb,amsfonts}
\usepackage{algorithmic}
\usepackage{graphicx}
\usepackage{float}
\usepackage{subfigure}
\usepackage{epsfig}
\usepackage{pifont}
\usepackage{textcomp}
\usepackage{xcolor}
\usepackage{subcaption}
\usepackage{algorithm}	
\usepackage{enumitem}
\usepackage{tikz}
\usepackage{listofitems} 
\usepackage{multicol}
\usepackage{etoolbox}
\usepackage{booktabs}
\usepackage{titlesec}
\usepackage{chngcntr}
\usepackage[inkscapeformat=png]{svg}
\usepackage{cite}
 \usepackage{color, soul} 
\usepackage{cleveref}

\usetikzlibrary{arrows.meta} 
\usepackage[outline]{contour} 
\contourlength{1.4pt}

\colorlet{myred}{red!80!black}
\colorlet{myblue}{blue!80!black}
\colorlet{mygreen}{green!60!black}
\colorlet{myorange}{orange!70!red!60!black}
\colorlet{mydarkred}{red!30!black}
\colorlet{mydarkblue}{blue!40!black}
\colorlet{mydarkgreen}{green!30!black}

\tikzset{
  >=latex, 
  node/.style={thick,circle,draw=myblue,minimum size=22,inner sep=0.5,outer sep=0.6},
  node in/.style={node,green!20!black,draw=mygreen!30!black,fill=mygreen!25},
  node hidden/.style={node,blue!20!black,draw=myblue!30!black,fill=myblue!20},
  node convol/.style={node,orange!20!black,draw=myorange!30!black,fill=myorange!20},
  node out/.style={node,red!20!black,draw=myred!30!black,fill=myred!20},
  connect/.style={thick,mydarkblue}, 
  connect arrow/.style={-{Latex[length=4,width=3.5]},thick,mydarkblue,shorten <=0.5,shorten >=1},
  node 1/.style={node in}, 
  node 2/.style={node hidden},
  node 3/.style={node out}
}
\def\nstyle{int(\lay<\Nnodlen?min(2,\lay):3)} 

\def\BibTeX{{\rm B\kern-.05em{\sc i\kern-.025em b}\kern-.08em
    T\kern-.1667em\lower.7ex\hbox{E}\kern-.125emX}}
\begin{document}

\title{BACSA: A Bias-Aware Client Selection Algorithm for Privacy-Preserving Federated Learning in Wireless Healthcare Networks}
\author{Sushilkumar~Yadav,~\IEEEmembership{Member,~IEEE,} and Irem~Bor-Yaliniz,~\IEEEmembership{Senior Member,~IEEE}
}


\maketitle

\begin{abstract}
Federated Learning (FL) has emerged as a transformative approach in healthcare, enabling collaborative model training across decentralized data sources while preserving user privacy. However, performance of FL rapidly degrades in practical scenarios due to the inherent bias in non Independent and Identically distributed (non-IID) data among participating clients, which poses significant challenges to model accuracy and generalization. Therefore, we propose the Bias-Aware Client Selection Algorithm (BACSA), which detects user bias and strategically selects clients based on their bias profiles. In addition, the proposed algorithm considers privacy preservation, fairness and constraints of wireless network environments, making it suitable for sensitive healthcare applications where Quality of Service (QoS), privacy and security are paramount. Our approach begins with a novel method for detecting user bias by analyzing model parameters and correlating them with the distribution of class-specific data samples. We then formulate a mixed-integer non-linear client selection problem leveraging the detected bias, alongside wireless network constraints, to optimize FL performance. We demonstrate that BACSA improves convergence and accuracy, compared to existing benchmarks, through evaluations on various data distributions, including Dirichlet and class-constrained scenarios. Additionally, we explore the trade-offs between accuracy, fairness, and network constraints, indicating the adaptability and robustness of BACSA to address diverse healthcare applications. 
\end{abstract}

\input{1_Introduction.tex}

\input{2_Preliminary.tex}
\input{3_Methodology.tex}
\input{4_Experiments.tex}

\section{Conclusions and Future Work}
\label{sec:Conc}
In this paper, we introduce BACSA to address the challenges of bias detection, client selection and privacy preservation in FL. BACSA reduces the bias introduced by non-IID data distributions across clients. This ensures that overrepresented classes do not disproportionately influence the model, while underrepresented classes are not neglected. As a result, BACSA promotes a more equitable learning process, ultimately leading to more balanced model performance across diverse client datasets, reducing bias, and enhancing fairness in federated learning scenarios compared to the state-of-art benchmarks. Furthermore, we proposed extension of BACSA with fixed sample size, BACSA-FS, and SNR consideration, BACSA-SNR. While BACSA-FS is shown to improve convergence graph, BACSA-SNR prioritizes channel conditions for client selection. To demonstrate robustness, we test BACSA on both widely used Dirichlet distribution and the proposed CCDD, which can represent challenging healthcare scenarios. Thanks to the proposed weight initialization and class estimation methods, similar accuracy to the methods in the literature with privacy violations are obtained while preserving the privacy of clients. For future work, we would like to support our observations and experiments with theoretical analysis and conduct experiments with data sets from healthcare. We also would like to investigate aspects of privacy protection beyond avoiding direct data or data characteristics sharing.

\section*{Acknowledgment}
Authors would like to thank Dr.~Gamini Senarath, Dr. Ngoc Dung Dao, Dr. Mohamed Alzenad, and Dr. Weisen Shi for their insightful discussions.

\bibliography{main.bbl}
\bibliographystyle{IEEEtran}

\end{document}

%% file: 1_Introduction.tex
\section{Introduction}
\begin{figure}[h!]
    \centering
    \includegraphics[width=0.9\linewidth]{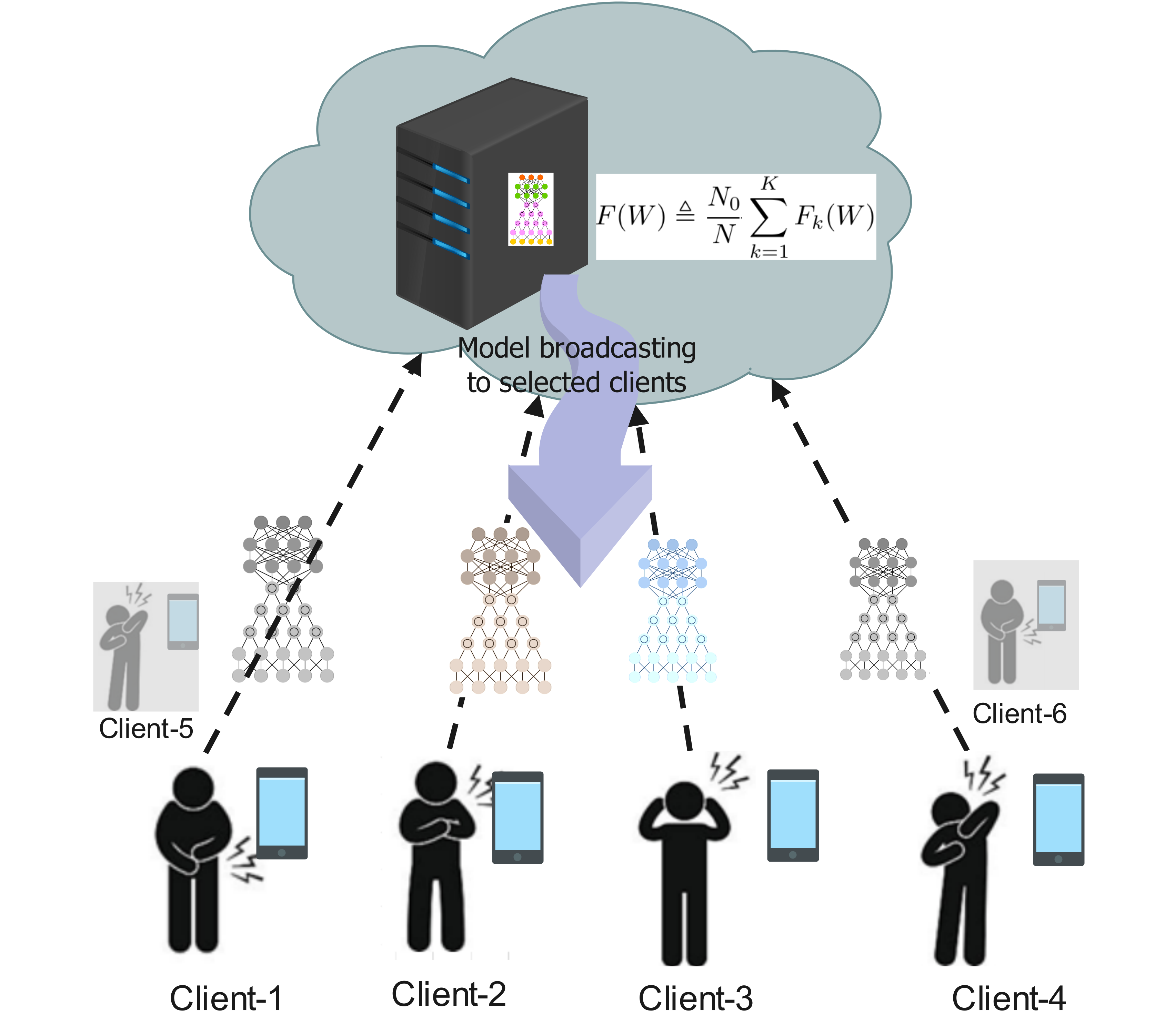}
    \caption{Illustration of FL in a wireless IoT for healthcare scenario: Clients have different health conditions causing severely non-IID data. A subset of clients need to be selected for each communication round due to system and device limitations: Clients 5 and 6 are omitted in this round.}
    \label{fig:system}
\end{figure}

Recent advancements on user equipment, sensors, processors and AI methods are transforming healthcare by generating and processing large volumes of granular data~\cite{health}. However, burdens of preserving the privacy of data and transferring local data to a centralized server to be processed creates many challenges. On the one hand, privacy-preserving distributed AI methods, such as Federated Learning (FL), hit two birds with one stone by allowing the participants to share the trained models instead of raw data, which are compact in size and protects data privacy~\cite{s_for_privacy}. On the other hand, FL provides comparable results to that of centralized machine learning methods solely under ideal conditions~\cite{s_fl_fundamental1, s_for_data_het_quantization, s_fl_fundamental}, and that is rarely the situation in practical applications. The participating devices would have non-homogeneous data distributions~\cite{s_data_het_2, s_for_data_het_quantization}, different computation capabilities~\cite{its_client_selection}, unstable communication channels and additional constraints due to their unique limitations, such as battery level. Among many practical considerations, the focus of this study is data heterogeneity, i.e., non-IID data distributions among clients. Non-IID data may be formed due to a variety of factors, including data size imbalance, for example when a hospital is data-rich whereas some clinics have scarce data, and/or data partitioning imbalance, where clients do not have complete data sets in terms of labels and/or features as illustrated in Fig.~\ref{fig:system} via clients with different health concerns~\cite{health, lee2020federated}. There are recent attempts to develop metrics and a through understanding of heterogeneity from the perspective of FL~\cite{data_metric}, however, none of the metrics in the literature are widely adopted by the community yet. 
Two benchmark studies in FL with non-IID distributions rely on random client selection~\cite{mcmahan_communication-efficient_2017, li_federated_2020}. On the contrary, recent studies in~\cite{chen_recovering_2024, wang_addressing_2020, yoshida2020_hybrid_fl, yang_federated_2020} employ various client selection (or model combination) strategies to mitigate adverse effects of non-IID data. However, these studies depend on the existence of auxiliary IID data, e.g., by collecting a subset of data from clients. The studies in \cite{astraea_duan, zhang_fed-cbs_2023} require users to share data characteristics, e.g., distribution of classes and number of samples, to select a statistically favorable client set. Both data collection in the former studies and exposure data characteristics in the latter ones are privacy violations. Instead, studies in~\cite{cluster_head} and \cite{its_client_selection} investigate model characteristics for client selection. For example, \cite{cluster_head} reduces communication cost and increases accuracy by clustering IoT users based on cosine difference among local models while considering strict networking constraints. However, the proposed architecture necessitates existence of devices that can serve as cluster heads and create new vulnerabilities against adversarial attacks. 
On the contrary to the limitations of previous studies, the proposed method can estimate the class distribution without additional data or architectural constraints by only investigating model parameters to reveal bias of clients with inspiration from the recent explainable-AI (XAI) studies. We then formulate the client selection problem as a mixed-integer non-linear optimization and use the bias estimations to obtain the most class balanced group while considering fairness among clients. We show that the proposed bias-aware client selection algorithm (BACSA) provides robustness and efficiency against non-IID data without exploiting or exposing clients, and therefore a suitable method to be used in healthcare applications.

The rest of the paper is organized as follows. Background information and notations are introduced in Sec.~\ref{sec:Pre}, followed by the presentation of the proposed method in Sec.~\ref{sec:Method}. Then we show the effectiveness of BACSA via experiments in Sec.~\ref{sec:Exp}, and conclude the paper with final discussions and future work in Sec.~\ref{sec:Conc}.

%% file: 2_Preliminary.tex
\section{Preliminaries}\label{sec:Pre}
The following is the terminalogy used regarding data sets~\cite{zhang_fed-cbs_2023}:
\begin{itemize}
	\item \textbf{Local dataset} means the local dataset owned by the clients. In this study, it is assumed that the local datasets are heterogenous both in terms of number of sample sizes and the distribution of the sample size over the classes. Therefore, the local data sets are both size imbalanced and class imbalanced. Class imbalance includes one or more data-rich classes, one or more classes with data scarcity and one or more absent classes for each client. 
	\item The aggregation of the local dataset from all the clients forms the \textbf{global dataset}. Global dataset is assumed to be class balanced, which means that all classes exist in the global set.
	\item \textbf{Grouped dataset} is the dataset formed after selecting the clients. The goal is to make the grouped data set as balanced as possible in terms of both class and size.
\end{itemize}

We assume there are $K$ clients of an FL scheme, and there is one central server to operate FL. Let $E(W)$ be the error function associated with the global objective function, $F(W)$ where $W$ indicates the model.  The central server has an initial model $W_0$, which must be distributed to each client, $k$, for them to train the model on their local data set $\mathbb{D}_k$. Once the trained initial model by a client $k$, ${W_0^k}$, is receieved, bias towards each class $\beta_ik$, can be estimated as the proportion of number of samples from each class as shown in Sec.~\ref{sec:bias}.
Assume a neural network with one input layer with $I+1$ input nodes, one output layer with $M+1$ nodes, and $S$ hidden layers with $L+1$ nodes. Also let $r$, $s$, and $u$ denote input layer, hidden layers, and output layer, respectively. Then, let $x_{v}^{(r)}$ represent a neuron $v$ at the $r^\text{th}$ input layer, and $\boldsymbol{\omega^{(m)}} = [{\omega^{(m)}_1, \omega{^{(m)}_2}, ..., \omega{^{(m)}_{M+1}}}]$ represent the vector of last layer weights. Let $\mathbb{D}_k=\{d_1, d_2, ...., d_j, ..., d_{N_k}\}$ be the data samples of client $k$, where $|{\mathbb{D}_k}|\triangleq N_k$ is the total number of samples of client $k$, and $j$ indicates data samples throughout this paper. Then, the total number of samples available globally is $N \triangleq \sum_k^K N_k$. The global set of classes is defined as $\mathbb{C}=\{c_1, c_2, ..., c_i\}$, where $|{\mathbb{C}_k}|\triangleq \Gamma$ indicates the total number of classes. Furthermore, $p_{ik}$ indicates true proportion of class $i$ in $\mathbb{D}_k$, as defined in Sec.~\ref{sec:bias}. For a class $c_i$, the true and predicted labels are indicated by  $y_i$ and $\hat{y_i}$, respectively. A binary variable $\lambda_k$ is used to indicate whether a client $k$ participates in the current communication round, $t$. 

%% file: 3_Methodology.tex
\section{Methodology}
\label{sec:Method}
%
%
%
In order to combat non-IID data with both class partition and data size disparities, multiple issues need to be addressed as discussed so far. Therefore, the proposed model has four parts:
\begin{itemize}
\item A specific weight initialization for $W_\text{0}$ which is shown to help with bias removal and increase accuracy compared to random weight initialization, 
\item The bias detection method which indicates the data-rich, scarce or absent classes in local data, 
\item Fixing the number of local samples that are used by each client to train the model, which prevents the bias of data-rich clients and/or classes, and
\item Client selection based on the detected bias to select a subset of clients whose models provides minimum bias after aggregation.
\end{itemize}
\subsection{Weight Initialization}
\label{SubSec: WeightInit}
Prior to training,  last layer weights of the initial global model, $W_\text{0}$ are initialized as follows.
\begin{equation}
\omega_{0} = \sqrt{\cfrac{1}{I_f \times L \times M}},
\label{eq:init}
\end{equation}
where $I_f$ is the number of input features, $M$ is the number of neurons in the output layer, and $L$ is the number of neurons in the second last layer. The typical value range in \eqref{eq:init} is lower than random values used in the state-of-the-art methods, such as Xavier-Glorot \cite{pmlr-v9-glorot10a} and He \cite{he_weight_init}, and ensures to capture the gradient change. In Sec.~\ref{sec:Exp}, we show the improvement in both bias detection and prediction accuracy with Monte Carlo simulations.

\subsection{Bias Detection}\label{sec:bias}

Let $\hat{y_i}$ be the predicted label and let $y_i$ be the true label for a particular class $i$, and $a_{j}^{(s)}$ indicate the activation output for a sample $d_j$. For a particular client $k$\footnote[1]{Index omitted for brevity in ~\eqref{eq:gradient} - ~\eqref{eq:binary_estimate}},
\begin{equation}
\label{eq:gradient}
\begin{aligned}
	& \frac{\partial E}{\partial {\omega}_{j}^{(u, s)}} =  (\hat{y}_{ij} -  y_i)\ {a}_{j}^{(s)}, \text{and}\\
      &\frac{\partial E}{\partial{w}_{j}^{(s, r)}} = ((\hat{y}_{ij} - y_i )\ {w}_{j}^{(s, r)} )\ \boldsymbol{{x}}_{j}^{(r)}\\
\end{aligned}
\end{equation} 
%
%
As a result, the gradient can be calculated as follows.
\begin{equation}
\label{eq:gradient1}
\begin{aligned}
	& \therefore \nabla E_{j} (\mathbf{W}) = \bigg[\frac{\partial E}{\partial \omega_{j}^{(u, s)}}, \frac{\partial E}{\partial w_{j}^{(s, r)}}\bigg],\\
	& \therefore \nabla E (\mathbf{W}) =\sum_{j=1}^{N_k} \nabla E_{j} (\mathbf{W}).
\end{aligned}
\end{equation} 
Then, the weight update expression becomes
\begin{equation}
\label{eq:weightupdate}
\begin{aligned}
	& \mathbf{W}_{\text{new}} = \mathbf{W}_{\text{old}} - \eta (\nabla E(\mathbf{W})),
\end{aligned}
\end{equation}
where $\eta$ is the learning factor. Furthermore, considering only the last layer
\begin{equation}
\label{eq:omega}
\begin{aligned}
	\omega_{j}^{u, s} &= \omega_{j}^{(u, s)} - \eta (\hat{y}_{ij} - y_j )\ a_{j}^{(s)}\\
	& = \omega_{j}^{(u, s)} - \eta \nabla (\boldsymbol{\omega}).
\end{aligned}
\end{equation}
In terms of existence of a class $i$ in $\mathbb{D}_k$,  \eqref{eq:omega} indicates
\begin{enumerate}
	\item If a client has data from class $i$, then $( \hat{y}_{ij} - y_i)$ will be non-positive, therefore $\nabla (\boldsymbol{\omega})$ will be negative. Then, $\omega^{(u,s)}_j$ will be non-negative.
	\item Otherwise, $( \hat{y}_{ij} - y_i)$ will be non-negative, therefore $\nabla (\boldsymbol{\omega})$ will be positive. Then, $\omega^{m,s}_j$ will be non-positive.
\end{enumerate}
As a result of~\eqref{eq:omega}, the global server can calculate the average of the weights for all the classes from the last layer of a received model, $W_k$. If a client $k$ does not have data from all the classes, the trained model's last layer will show bias, in the form of dominant weights for the existing and/or data-rich classes in $\mathbb{D}_k$.  As stated in~\cite{anand_binary_estimation}, when training DNN for a binary classification problem consisting of label $a$ and label $b$ respectively, the expectations of gradient square for different classes have the following approximate relation
\begin{equation}
\label{eq:binary_estimate}
\frac{\mathbb{E}\|\nabla E (\boldsymbol{\omega}_{a})\|^2}{\mathbb{E}\|\nabla E (\boldsymbol{\omega}_{b})\|^2}  \approx \frac{n_a^2} {n_b^2},
\end{equation} 
where $E$ denotes the error function of the neural network, $n_a\ \text{and}\ n_b$ are the number of samples for class $a$ and class $b$, respectively. 

In this study, we extend the above idea to a multi-class scenario. Initialize the last layer weights with the initialization mentioned in Sec.~\ref{SubSec: WeightInit}. After training, it satisfies the following relation\\
	\begin{equation}
\label{eq:prop}
	\begin{aligned}
	& \frac{\mathbb{E}\|max(0, \boldsymbol{\omega}^{new}_{i, k})\|^2}{\sum_{c=1}^{\Gamma}\mathbb{E}\|max(0, \boldsymbol{\omega}^{new}_{c, k} )\|^2} \approx \frac{n_{i, k}^2}{\sum_{c=1}^{\Gamma} n_{l, k}^2},\\
	\end{aligned}
	\end{equation}
where $n_i\ \text{and}\ n_l$ are the number of samples for class $i$ and class $l$, respectively, where $i \neq l,\ c \in \mathbb{C}_k$.
To bring the ratio on a common scale for all the clients, we modify the ratio as follows
	\begin{equation}
		\mathbf{\beta}_{i, k} = \frac{\mathbb{E}\|max(0, \boldsymbol{\omega}^{new}_{i, k})\|^2} {\sum_{k =1}^{K} \sum_{c=1}^{\mathbb{C}_k}\mathbb{E}\|max(0, \boldsymbol{\omega}^{new}_{l, k} )\|^2} \approx \frac{n_{i, k}^2}{\sum_{k = 1}^{K}\sum_{c=1}^{\Gamma}n_{l,k}^2},
\label{eq:beta}
	\end{equation}
where $\beta_{i, k}$ be the estimated proportions for class $i$ in client $k$. Then, the percentage error in estimating the proportion for a class $i$, $\kappa_i$, is obtained as
\begin{equation}
\label{eq:percentage}
	\kappa_i = \frac{1}{K}\sum_{k=1}^{K} \frac{|p_{ik} - \beta_{ik}|}{p_{ik}} \times 100.
\end{equation}

\subsection{Bias-Aware Client Selection}\label{AA}
In this section, we propose the bias-aware client selection algorithm (BACSA), which is based on the formulation of the problem as a mixed integer non-linear problem.

Assume $ \lambda_k $ be the binary variable that indicates whether a client $k$ is selected or not and  $\beta_{i, k}$  be the estimated proportion of class $i$ in $\mathbb{D}_k$. Let S/N be the Signal-to-Noise ratio and Let $\overline{q_i}$ be the average proportion of class $i$ across the selected clients such that
 
\begin{equation}
\begin{aligned}
       \overline{q_i} =  \sum_{i=1}^{\Gamma} \sum_{k=1}^K 1^N \lambda_k \beta_{i, k}, \\
	\text{and then variance of } q_i\ \text{becomes}\\
	\text{Var}(\mathbf{q_i}) = \sum_{i=1}^{\Gamma} \sum_{k=1}^K \lambda_k (\beta_{i, k} - \overline{q_i})^2\\
\end{aligned}
\end{equation}
Then the objective function is to minimize the variance to encourage uniformity in the selected clients becomes
\begin{equation}
        \operatorname*{min}_{q} { \sum_{i=1}^{\Gamma}   \sum_{k=1}^K \lambda_k \bigg ( (\beta_{i, k} - \overline{y_q})^2} + \gamma \  \sqrt{\frac{3 \  ln(r) * 2\ m_{tk}}{\theta *  \text{S/N}}}\bigg),
\label{eq:SNR}
\end{equation}
where $\Gamma$ is the total number of classes, $m_{tk}$ is the count of client  $k$ at round  $r$, $\gamma$ is the exploration and exploitation factor, $\theta$ is a scaling/weighting factor. Consider the following distributed optimization over FL, when the number of local training samples that can be used is fixed as $N_0$
\begin{equation}
\min_{W \in \mathbb{R}^{d} }\bigg\{ F(W) \triangleq \frac{N_0}{N} \sum_{k=1}^K {F_k(W)}\bigg\},
\end{equation}
Then the typical weighted term for the $k^{th}$ device can be omitted and the global objective function, $F(W)$ can be written as
\begin{equation}
\min_{W \in \mathbb{R}^{d} }\bigg \{ F(W) \triangleq \sum_{k=1}^K {F_k(W)}\bigg\}.
\end{equation}
Since ${F_k(W)}$ represents the expected loss of the user $k$, The local objective function ${F_k(w)}$ is defined as 
\begin{equation}
{F_k(W)} \triangleq \cfrac{1}{N_0}\sum_{j=1}^{N_0} (l_k(W),
\end{equation}
$l_k(W)$ measures the loss of model $W$ in predicting label $y$ with input $x$. BACSA is summed in~\ref{algorithm1}.

\begin{algorithm}
    \caption{Bias-Aware Client Selection Algorithm (BACSA)}\label{algorithm1}
    \begin{algorithmic}
        \STATE \textbf{Input: }  $\mathrm{Initialize\ Global\ Model\ F(W)}$ with weights $W_0$, Fixed sample size $N_0$, training sample  $\xi$, 
	 \STATE \textbf{Output: } Trained global model 
        \FOR{t = $1$ to $T$ do}
		\IF{t = 1, 2, ...., till each client $k$ is covered once}
			\FOR {$ \text{all }k \in K \textbf{in parallel}$}
				\FOR {$e = 0, 1,..., \varepsilon-1$} 
	                \STATE $W_{t+e+1}^k \gets W_{t+e}^k - \eta_{t+e} \Delta F_k (W_{t+e}^k , \xi_{t+e}^k)$
				\ENDFOR
	                \STATE upload $W_{t}^k$ to the server
	            \ENDFOR
			\STATE estimate $\beta_{i, k}$ for all clients $k$ using~\eqref{eq:prop}
		\ELSE
			\STATE Select optimum subset of clients $S_k$ which will form the global balanced set and has minimal variance amongst the classes
			\FOR{$ \text{all}\ k \in K \ \textbf{in parallel}$}
			\FOR {$e = 0, 1,..., \varepsilon-1$} 
	                \STATE $W_{t+e+1}^k \gets W_{t+e}^k - \eta_{t+e} \Delta F_k (W_{t+e}^k , \xi_{t+e}^k)$
			\ENDFOR
	                \STATE upload $w_{t}^k$ to the server
	            \ENDFOR
		\STATE $\min_{W \in \mathbb{R}^{d} }\{ F(W) \triangleq \sum_{k=1}^K {F_k(W)}\}$
		\ENDIF
            
        \ENDFOR
    \end{algorithmic}
\end{algorithm}
\begin{figure*}[hbtp]
  \centering
  \subfigure[]{\includegraphics[width=0.24\textwidth]{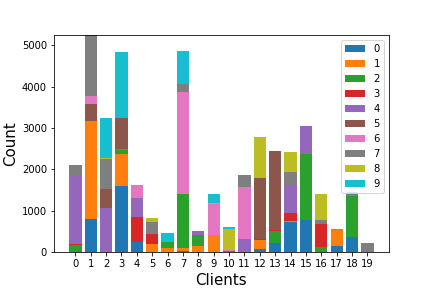}\label{d1}}
  \subfigure[]{\includegraphics[width=0.24\textwidth]{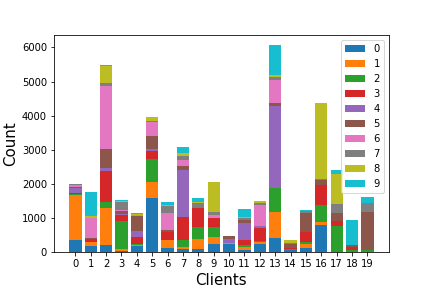}\label{d2}}
   \subfigure[]{\includegraphics[width=0.24\textwidth]{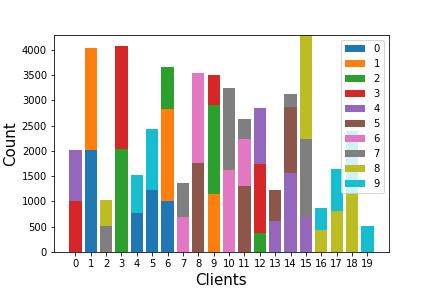}\label{c1}}
  \subfigure[]{\includegraphics[width=0.24\textwidth]{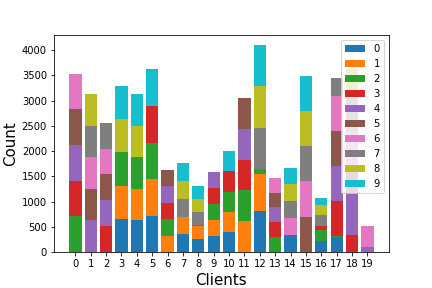}\label{c2}}
 \caption{Data distributions based on Dirichlet and CCDD using CIFAR-10 dataset. (a) Dirichlet distribution with $\alpha = 0.1$. (b) Dirichlet distribution with $\alpha = 1$. (c) CCDD with  $\Phi = 2$ and $\Gamma = 10$. (d) CCDD with  $\Phi = 5$ and $\Gamma = 10$. }
  \label{fig:generic}
\end{figure*}

%% file: 4_Experiments.tex
\section{Experiments and Results}\label{sec:Exp}
\begin{figure}[h]
    \centering
    \begin{minipage}{0.3\linewidth} 
        \centering
        \begin{tabular}{|c|p{2.5cm}|} 
            \hline
            \textbf{Number} & \textbf{Label} \\ \hline
            0  & Atelectasis \\ \hline
            1  & Cardiomegaly \\ \hline
            2  & Effusion \\ \hline
            3  & Infiltration \\ \hline
            4  & Mass \\ \hline
            5  & Nodule \\ \hline
            6  & Pneumonia \\ \hline
            7  & Pneumothorax \\ \hline
            8  & Consolidation \\ \hline
            9  & Edema \\ \hline
            10 & Emphysema \\ \hline
            11 & Fibrosis \\ \hline
            12 & Pleural Thickening \\ \hline
            13 & Hernia \\ \hline
        \end{tabular}
        \label{nihlabels}
    \end{minipage}%
    \hfill
    \begin{minipage}{0.47\linewidth} 
        \centering
        \includegraphics[width=\linewidth]{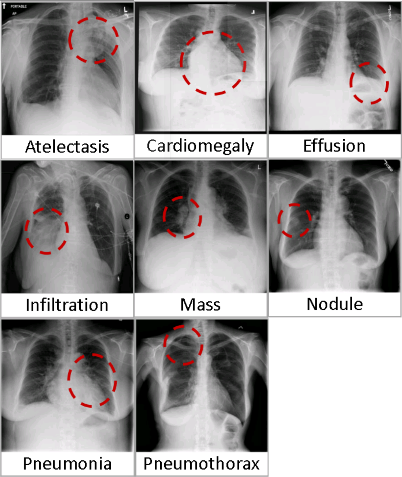}
        \label{nihimage}
    \end{minipage}
     \caption{NIH-CXR dataset \protect{\cite{wang2017hospital}}}
    \label{nihdataset}
\end{figure}

In all our experiments, we simulated cross device federated learning where only a fraction of devices will be available in each communication cycle. Since most of the devices are resource constrained, we used a CNN model with two convolution layers and three Feed Forward Neural Network (FFNN) layers. Each device performs the computation for 5 epochs with a batch size of 32. Under ideal conditions, the benchmark accuracy with this model is approximately $\sim$ 84\%  with CIFAR-10 dataset \cite{krizhevsky_cifar10} consisting of 60000 32x32 colour images in 10 classes, with 6000 images per class. There are 50000 training images and 10000 test images. The local optimizer is SGD with a weight decay of $5\mathrm{e}{-4}$, and learning rate, $\eta = 0.01$. We terminate the FL training after 2000 communication rounds and then evaluate the model’s performance on the test dataset of CIFAR-10. In addition to CIFAR-10, we also used the NIH Chest X-ray (CXR) dataset \cite{wang2017hospital} to demonstrate our class proportion estimation mechanism. NIH dataset consists of 51,759 samples from 14 classes as shown in Fig.~\ref{nihdataset}, excluding the `No Finding' class with each sample size of 1024x1024 pixels.

\begin{figure*}[h!]
  \centering
  \subfigure[]{\includegraphics[width=0.75\textwidth]{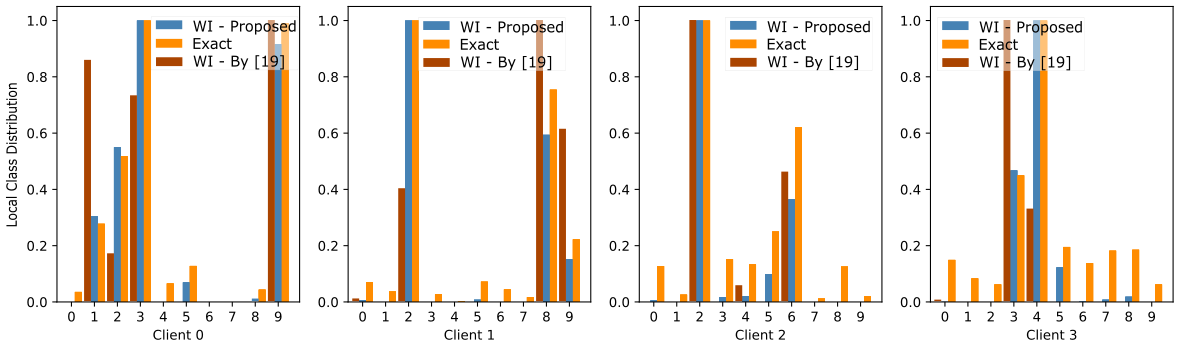}\label{labelestimationindividualdirichlet}}
  \subfigure[]{\includegraphics[width=0.75\textwidth]{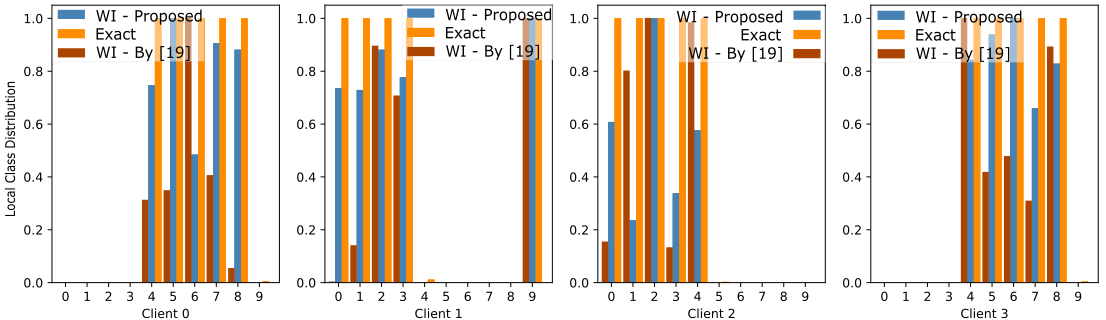}\label{labelestimationindividuaccddl}}
  \subfigure[]{\includegraphics[width=0.75\textwidth]{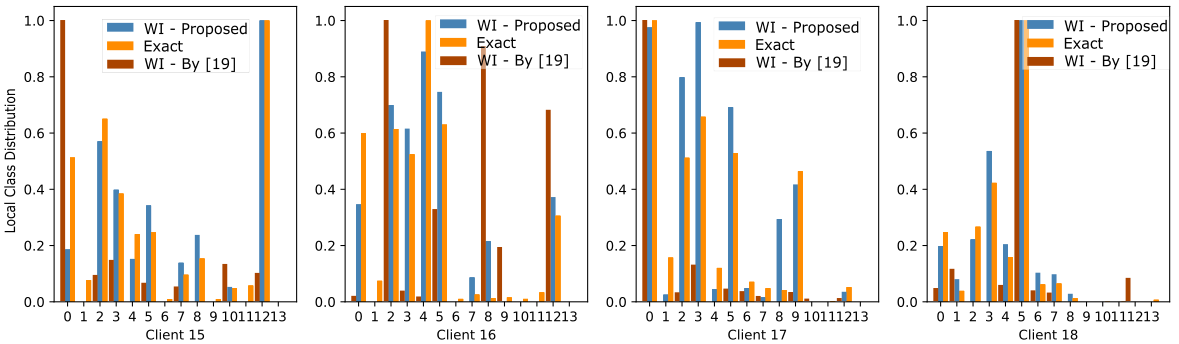}\label{labelestimationindividualnihdirichlet}}
 \caption{Class estimation accuracy using proposed weight initialization (WI) and WI as mentioned in \cite{pmlr-v9-glorot10a} (a) Dirichlet $\alpha = 0.5$. (b) CCDD $\Phi = 5$. (c) NIH-CXR Dirichlet $\alpha = 0.2$.}
  \label{fig:generic}
\end{figure*}

We use IID data distribution as the benchmark, where the total number of samples at each client is equal ($N_k = N, \forall k$), and each client $k$ has equal number of samples from all classes. For non-IID scenario, the global data set contains equal samples of all the classes, however, that is distributed unevenly to local clients. We considered two strategies for data distribution:
\begin{enumerate}
\item Dirichlet-$\alpha$ distribution \cite{hsu_federated_2020} is parameterized by the concentration parameter, $\alpha$. As $\alpha$ decreases, the data distribution becomes more heterogenous as seen in Fig.~\ref{d1} and~\ref{d2}. It is used as a benchmark from the literature.
\item In order to further test BACSA, we utilize a class-constrained data distribution (CCDD), where each client can have only a certain number of classes, $\Phi$. This scenario demonstrates the effect of missing class as seen in Fig.~\ref{c1} and~\ref{c2}.
\end{enumerate}
The communication channel is assumed to be stable across all clients and all clients can send their trained model to the global model. At a particular communication cycle, only a subset of clients will participate in the FL process. FL setup is same throughout all our experiments. Distribution of medical data between hospitals is demonstrated by using MNIST dataset in \cite{lee2020federated}. Similarly, we utilize the CIFAR-10 dataset, which consists of images more closely resembling medical images in terms of complexity compared to MNIST dataset. We assume a total of 20 clients (hospitals) to best simulate a real-world environment and the client selection algorithm selects 5 of them in each communication cycle.

\begin{figure}[h!]
    \centering
    \begin{minipage}{\linewidth}
        \centering
        \includegraphics[width=.8\linewidth]{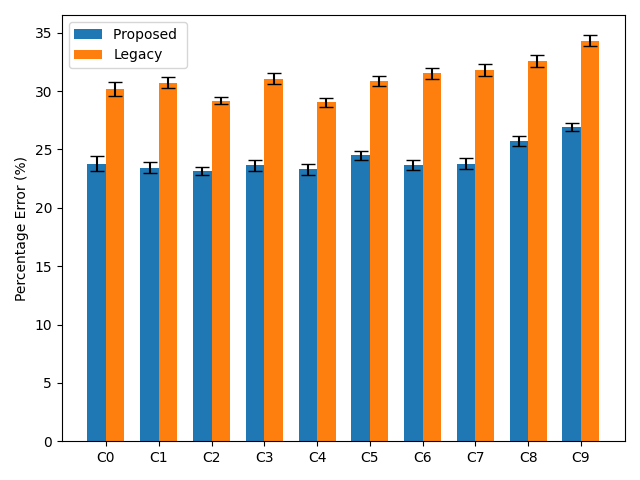}
        \caption{Comparing legacy and proposed weight initialization methods on class estimation accuracy using CIFAR-10 dataset}
        \label{montecarlo} 
    \end{minipage}
\end{figure}

Fig.~\ref{montecarlo} demonstrates the effect of weight initialization proposed in~\eqref{eq:init} compared to a widely used legacy initialization \cite{pmlr-v9-glorot10a}. We use 100 Monte Carlo simulations  $H = 100$  with Dirichlet distribution ($\alpha=0.1$). In order to calculate estimation error, we use~\eqref{eq:percentage} such that $\kappa^H_i = \sum_{h = 1}^{H}{\kappa_i^h}$, where $h$ indicates each Monte Carlo simulation. Results show a consistently superior performance with up to 12\% more accuracy in class proportion estimation. 
For Dirichlet-based non-IID data distribution with $\alpha = 0.5$, Fig.~\ref{labelestimationindividualdirichlet} shows the plot for class proportion estimation for individual clients using~\eqref{eq:beta}. For CCDD, Fig.~\ref{labelestimationindividuaccddl} shows the plot for label proportions estimation for individual clients using the weight ratios. For NIH Chest Xray dataset, Fig.~\ref{labelestimationindividualnihdirichlet} shows the plot for estimation of label proportions corresponding to each client by using weight ratios. It is clearly seen that the class proportion estimation using the proposed weight initialization is closer to the true class proportion compared to the legacy weight initialization. The proposed technique effectively captures underlying trends in the data distribution, allowing the model to identify bias and tackle for class imbalances more accurately.
%
%
Since CIFAR-10 is a global balanced dataset, when all clients [hospitals] participate in training, the final aggregated model does not diverge significantly, resulting in faster and smoother convergence. This can be used as a benchmark to evaluate the accuracy achieved on non-IID data.  However, if the dataset is not globally balanced, then the accuracy drops significantly \cite{ zhang_fed-cbs_2023} and this is where the bias removal algorithms need to be implemented. Bias reduction techniques become crucial in such scenarios to ensure that the model does not favor the over-represented classes while under-performing on the under-represented ones.

\begin{figure}[h!]
    \centering
    \includegraphics[width=0.9\linewidth]{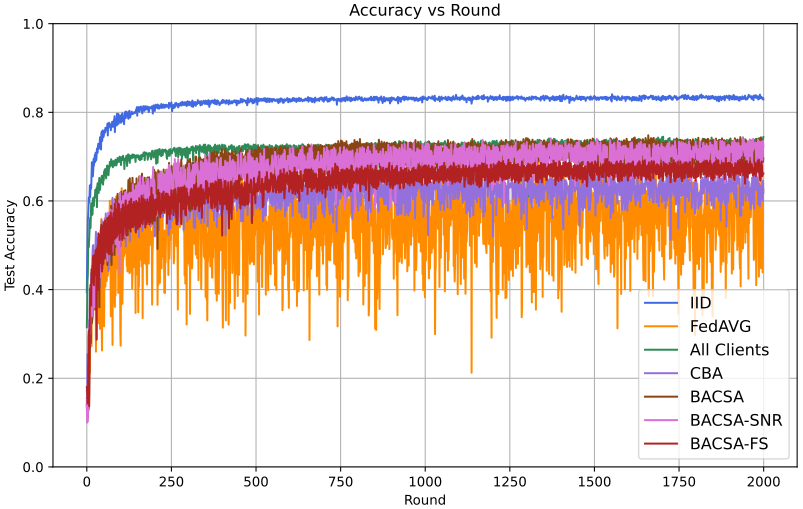}
    \caption{Accuracy results}
    \label{fig:accuracyresults}
\end{figure}

Fig.~\ref{fig:accuracyresults} and Table~\ref{tab:accuracyresults} demonstrates the performance of BACSA and BACSA with fixed sample size (BACSA-FS) compared to several benchmarks.  We also compared the result with client selection using greedy technique where it just selects clients which forms the most globally class balanced scenario. Results demonstrate improvement in the convergence and accuracy using BACSA as compared to the greedy technique.  

\begin{figure*}[h!]
  \centering
  \subfigure[]{\includegraphics[width=0.31\textwidth]{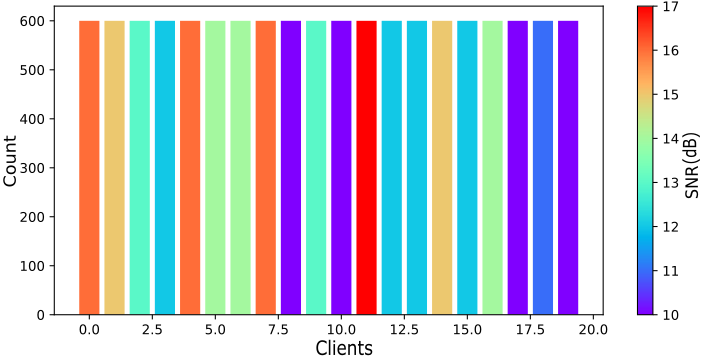}\label{fig:nosnrcount}}
  \subfigure[]{\includegraphics[width=0.31\textwidth]{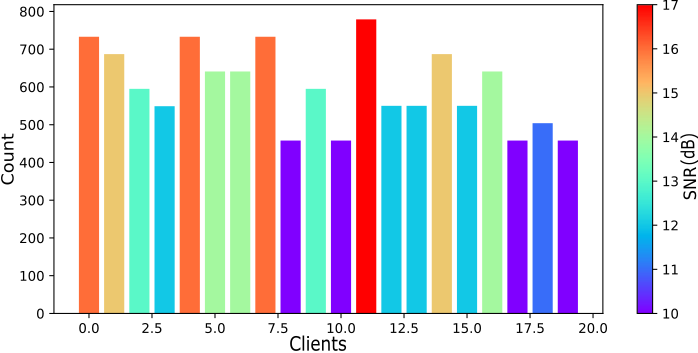}\label{fig:noweightcount}}
  \subfigure[]{\includegraphics[width=0.31\textwidth]{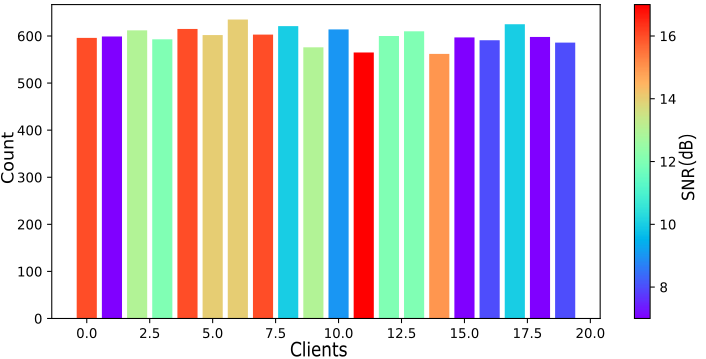}\label{fig:greedycount}}
 \caption{Fairness is demonstrated in terms of total times a client is request to participate in FL, $m_{k}$. BACSA and BACSA-SNR affect client selection $\theta$ in~\eqref{eq:SNR} and prevent exploitation of clients with favorable data characteristics.}
  \label{fig:count}
\end{figure*}
Using our proposed method, we can see that the convergence gD1raph improved significantly for the same non-IID (CCDD $\Phi=2$) data. BACSA-FS and BACSA achieves more stable convergence as seen in Fig.~\ref{fig:accuracyresults} and Table \ref{tab:accuracyresults}. The enhancement in stability can also be perceived as a reduction in gradient variance, a concept that has been explored in previous studies \cite{zhao_federated_2022}.
\begin{table}
    \centering
    \begin{tabular}{lcc}
        \toprule
        & \textbf{Accuracy (\%)} \\
        \midrule
        IID & 83.33 $\pm$ 0.25 \\
        FedAVG \cite{mcmahan_communication-efficient_2017} & 57.62 $\pm$ 15 \\
        All clients & 73.17 $\pm$ 0.75 \\
        CBA \cite{yang_federated_2020}  & 61.91 $\pm$ 5 \\
        CBA-FS  & 61.44 $\pm$ 2 \\
        BACSA & 70.05 $\pm$ 3 \\
        BACSA-FS & 68 $\pm$ 2 \\
	  BACSA-SNR & 67 $\pm$ 6 \\ 
        \bottomrule
    \end{tabular}
    \caption{Accuracy results for different methods}
    \label{tab:accuracyresults}
\end{table}

Furthermore, we demonstrate how BACSA provides fairness among FL clients, while improving accuracy in Fig.~\ref{fig:count}. Note that, without the $m_{tk}$ in~\eqref{eq:SNR}, BACSA would simply determine a subset of users with the most homogeneous combined data distribution based on class proportions. However, results in Fig.~\ref{fig:nosnrcount} show that $m_{tk}$ prevents exploitation of clients with desirable data characteristics. In addition, Fig.~\ref{fig:noweightcount} shows that BACSA-SNR increases participation of users with high Signal-to-Noise-Ratio (SNR), which increases efficiency of communications. On the other hand, due to their favorable data characteristics, some clients with low SNR participated more often than clients with high SNR in Fig.~\ref{fig:greedycount}, where Greedy approach is employed. 

Even though BACSA aims to select clients to minimize bias, the aggregation can still lead to some weight divergence. This is because the bias estimations are not exact and may include inaccuracies, which can contribute to the divergence. Nevertheless, by employing our technique, we can still achieve performance comparable to full client participation with a globally balanced dataset. Importantly, BACSA achieves this without violating individual privacy, as all data remains decentralized, and only model updates are shared during the aggregation process. This ensures that sensitive medical data from different clients (e.g., hospitals) is kept private while still benefiting from collective training.

For the NIH Health Care dataset, we demonstrated the bias estimation, which is the main focus of this paper. However, due to the time constraints, we were unable to train the CNN model to plot its accuracy. The image samples are 1024x1024 pixels, and even with resizing, training a larger CNN model would require additional time.